\documentclass[letterpaper]{article} 
\usepackage{aaai2026}  
\usepackage{times}  
\usepackage{helvet}  
\usepackage{courier}  
\usepackage[hyphens]{url}  
\usepackage{graphicx} 
\usepackage{natbib}  
\usepackage{caption} 
\usepackage{algorithm}
\usepackage{algorithmic}
\usepackage{amsmath}
\usepackage{multirow}
\usepackage{booktabs}
\usepackage{newfloat}
\usepackage{listings}
\DeclareCaptionStyle{ruled}{labelfont=normalfont,labelsep=colon,strut=off} 
\floatstyle{ruled}
\newfloat{listing}{tb}{lst}{}
\floatname{listing}{Listing}
\title{Guided Navigation in Knowledge-Dense Environments: Structured Semantic Exploration with Guidance Graphs}
\author{
    Dehao Tao\textsuperscript{\rm 1},
    Guangjie Liu\textsuperscript{\rm 2},
    weizheng\textsuperscript{\rm 1},
    Yongfeng Huang\textsuperscript{\rm 1},
    Minghu jiang\textsuperscript{\rm 1}  
}
\affiliations{
    
    \textsuperscript{\rm 1}	Tsinghua University\\
    \textsuperscript{\rm 2}	Xinjang University\\

}

\begin{document}

\maketitle

\begin{abstract}
While Large Language Models (LLMs) exhibit strong linguistic capabilities, their reliance on static knowledge and opaque reasoning processes limits their performance in knowledge-intensive tasks. Knowledge graphs (KGs) offer a promising solution, but current exploration methods face a fundamental trade-off: question-guided approaches incur redundant exploration due to granularity mismatches, while clue-guided methods fail to effectively leverage contextual information for complex scenarios.
To address these limitations, we propose Guidance-Graph-guided Knowledge Exploration (GG-Explore), a novel framework that introduces an intermediate Guidance Graph to bridge unstructured queries and structured knowledge retrieval. The Guidance Graph defines the retrieval space by abstracting the target knowledge’s structure while preserving broader semantic context, enabling precise and efficient exploration.
Building upon the Guidance Graph, we develop: (1) Structural Alignment that filters incompatible candidates without LLM overhead, and (2) Context-Aware Pruning that enforces semantic consistency with graph constraints.
Extensive experiments show our method achieves superior efficiency and outperforms SOTA, especially on complex tasks, while maintaining strong performance with smaller LLMs, demonstrating practical value.
\end{abstract}


\section{Introduction}
Large Language Models (LLMs) have demonstrated remarkable capabilities across diverse natural language tasks, including question answering \cite{wang2024t,li2024unigen,zhao2024comi}, text generation \cite{ji2024chain,chen2023table,chen2022entity,gong2024graph}, and recommender systems \cite{zhang2023triple,zhang2024temporal,wu2024afdgcf,wang2024dynamic}. By leveraging deep learning and vast training data, they achieve human-like language fluency. However, their knowledge is static post-training, limiting real-time updates; they may generate plausible but incorrect responses ("hallucinations") \cite{bang2023multitask,ji2023survey,luo2023augmented}. Integrating external knowledge sources (e.g., knowledge graphs) offers a promising solution \cite{zhang2019ernie,yao2019kg,wang2021kepler,luo2023reasoning}, underscoring the need for further research in this evolving field.

\begin{figure}[htb]
    \centering
    \includegraphics[width=1\linewidth]{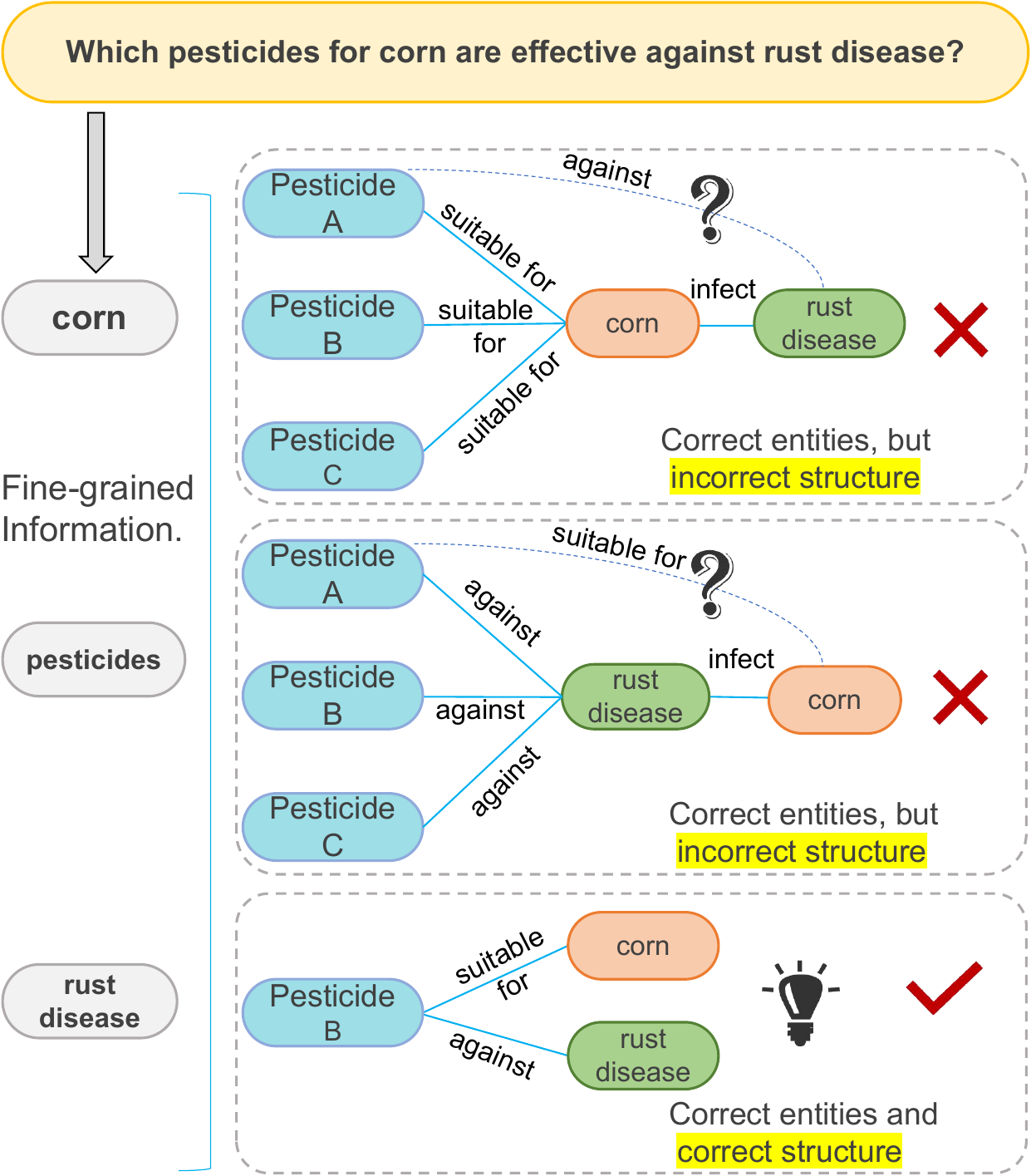}
    \caption{Relying solely on fine-grained information may fail to differentiate paths containing identical entities but distinct relational structures.}
    \label{fig:fg}
\end{figure}

Some researchers have attempted to directly employ LLMs to transform questions into database query statements for retrieving relevant knowledge from knowledge graphs \cite{hu2023chatdb,wang2023knowledgpt}. While this approach is straightforward, the generated queries may not always align well with the structure of the knowledge graph. A viable alternative provides KG context to the LLM, using question-guided multi-round iteration to progressively identify relevant knowledge \cite{li2023chain,zhao2023verify,sun2024think}. The question-guided approach fundamentally suffers from a granularity mismatch between natural language queries and KG entities, leading to two critical limitations: (1) imprecise knowledge targeting due to semantic misalignment, and (2) computational inefficiency from redundant graph exploration.

FiSKE extracts fine-grained keywords from queries for stateful knowledge retrieval, avoiding redundancy in stateless approaches \cite{TAO2025114011}. However, FiSKE discards other contextual information in the question to extract fine-grained keywords. While this simplification proves effective for structurally simple scenarios, it becomes inadequate when handling the intricate structural relationships inherent in knowledge-intensive contexts, as illustrated in the Figure \ref{fig:fg}.

In this paper, we propose Guidance-Graph-guided Knowledge Exploration (GG-Explore), a novel paradigm that bridges the gap between unstructured queries and structured knowledge retrieval. The novel paradigm introduces an intermediate Guidance Graph construction phase. The Guidance Graph defines the retrieval space, requiring all extracted knowledge to align with its structural and semantic framework. It abstracts the target knowledge's structure while preserving wider semantics, thereby directing retrieval with precision. 
Unlike existing methods that perform pruning solely based on relations and entities, our approach leverages the structural and contextual information provided by the Guidance Graph to design structural alignment and context-aware pruning mechanisms. These components selectively filter and match target knowledge from both structural and semantic perspectives. Specifically, structural alignment efficiently eliminates candidates structurally inconsistent with the Guidance Graph without requiring model computations, while context-aware pruning precisely identifies knowledge that aligns with the contextual information in the Guidance Graph.

The contributions of this paper are as follows:
\begin{enumerate}
    \item We propose a novel question-Guidance Graph-knowledge paradigm that introduces an intermediate Guidance Graph construction phase to bridge unstructured queries and structured knowledge retrieval. The Guidance Graph defines the retrieval space by abstracting the target knowledge's structure while preserving broader semantics.

    \item We develop two core mechanisms based on the Guidance Graph: structural alignment and context-aware pruning. These components work synergistically to precisely filter and match target knowledge from both structural and semantic perspectives, significantly improving retrieval efficiency and accuracy.

    \item Extensive experiments demonstrate the effectiveness of our method, with three key advantages: (1) significantly higher answer accuracy than existing methods, particularly for complex questions; (2) high efficiency with minimal LLM calls and token usage; (3) outstanding performance on small-parameter LLMs, highlighting strong practical value.
\end{enumerate}

\begin{figure}[htb]
    \centering
    \includegraphics[width=0.95\linewidth]{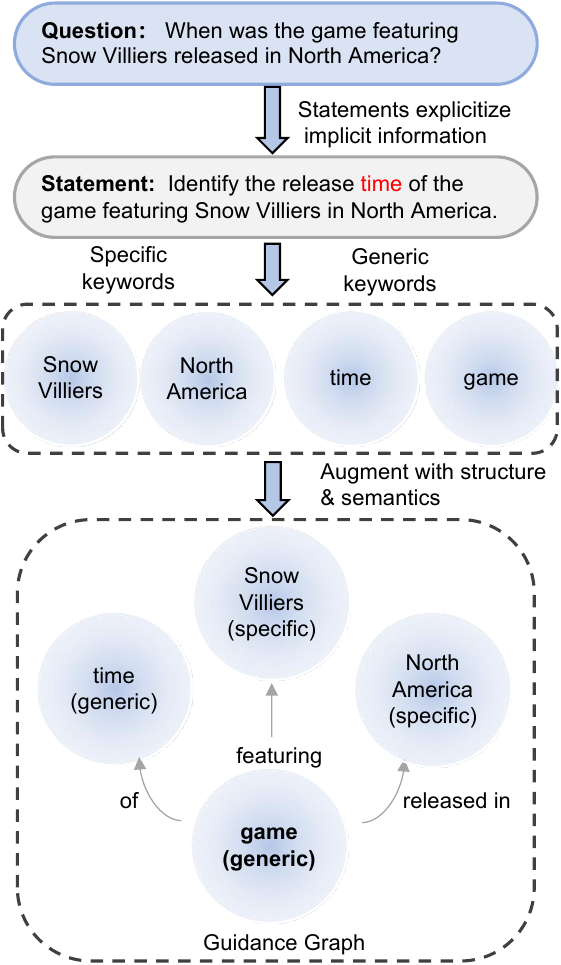}
    \caption{Workflow of Guidance Graph Construction}
    \label{fig:GMEG}
\end{figure}

\section{Related work}
\subsection{LLM Resoning with Prompt}
Recent advancements in the reasoning capabilities of LLMs have been driven by various prompting techniques aimed at improving their performance on complex tasks. DecomP~\cite{a6} breaks down reasoning tasks into manageable sub-tasks, allowing for step-by-step problem-solving. Chain-of-Thought (CoT) ~\cite{a1} and its derivatives, such as Tree-of-Thought(ToT)~\cite{a2}, Graph-of-Thought(GoT)~\cite{a3}, and Memory of Thought(MoT)~\cite{a4}, have been instrumental in encouraging LLMs to generate intermediate reasoning steps, thereby enhancing their cognitive processes. Plan-and-solve~\cite{a5} prompts LLMs to formulate plans and conduct reasoning based on these plans.  

\subsection{Question Answering with KG-Augmentd LLM}
Knowledge Graph Question Answering (KGQA) has evolved significantly with the integration of LLMs and Knowledge Graphs. 
Early approaches embedded KG knowledge into LLMs during pre-training or fine-tuning~\cite{a18}, but this limited explainability and update efficiency. To enhance reasoning, retrieval-augmented techniques retrieve relevant facts from KGs \cite{a14}. UniKGQA~\cite{a15} unifies graph retrieval and reasoning within a single LLM model, achieving state-of-the-art results. Recent methods translate KG knowledge into textual prompts for LLMs, enhancing reasoning without sacrificing KG's strengths~\cite{a19}. For instance, generating SPARQL queries or sampling relevant triples aids LLM inference~\cite{a20}. The KG-augmented LLM paradigm treats LLMs as agents exploring KGs interactively, improving reasoning capabilities~\cite{a21}.

\section{Method}
\subsection{Problem Definition}
Knowledge Graph Question Answering involves querying structured knowledge graphs to generate accurate responses to natural language questions. 
In our proposed method, we first extract information from the question to construct a Guidance Graph $GG = (e,v)$. Subsequently, we identify all mappings of nodes $clue\_e$ and edges $clue\_r$ from $GG$ to entities and relations from knowledge graph during the knowledge exploration process. Finally, we retrieve the corresponding knowledge from these mappings and provide it to the LLM for question answering.

\subsection{Guidance Graph Construction}
\label{GMEG construction}
Our framework constructs Guidance Graphs as illustrated in Figure \ref{fig:GMEG}.

We employ LLMs to perform hierarchical processing of questions. First, we transform elliptical questions into complete declarative statements to resolve information incompleteness.
Second, we extract fine-grained elements from these statements, categorizing them into: (1) specific keywords (named entities like `North America') and (2) generic keywords (broader terms like "country").
Third, we mine semantic relationships between keywords to reinforce the structural representation of the Guidance Graph

To establish the graph structure while enriching semantic information, we design the following generation rules based on specific and generic keywords:

\begin{enumerate}
    \item Specific keywords exclusively serve as graph nodes (i.e., entities in triples) rather than edges.
    \item When a generic keyword co-refers to the same entity as a specific keyword, it functions as a triple's relation rather than an entity.
    \item For two distinct generic keywords referencing the same entity, one must be assigned as the relation in the triple.
    \item For associated keywords referring to different entities, we construct triples where the keywords become head/tail entities and their association forms the relational edge. 
\end{enumerate}

These rules ensure that the structure and semantics of the Guidance Graph closely align with the original intent of the question.
The prompts used are shown in Appendix.

\begin{figure*}
    \centering
    \includegraphics[width=0.95\linewidth]{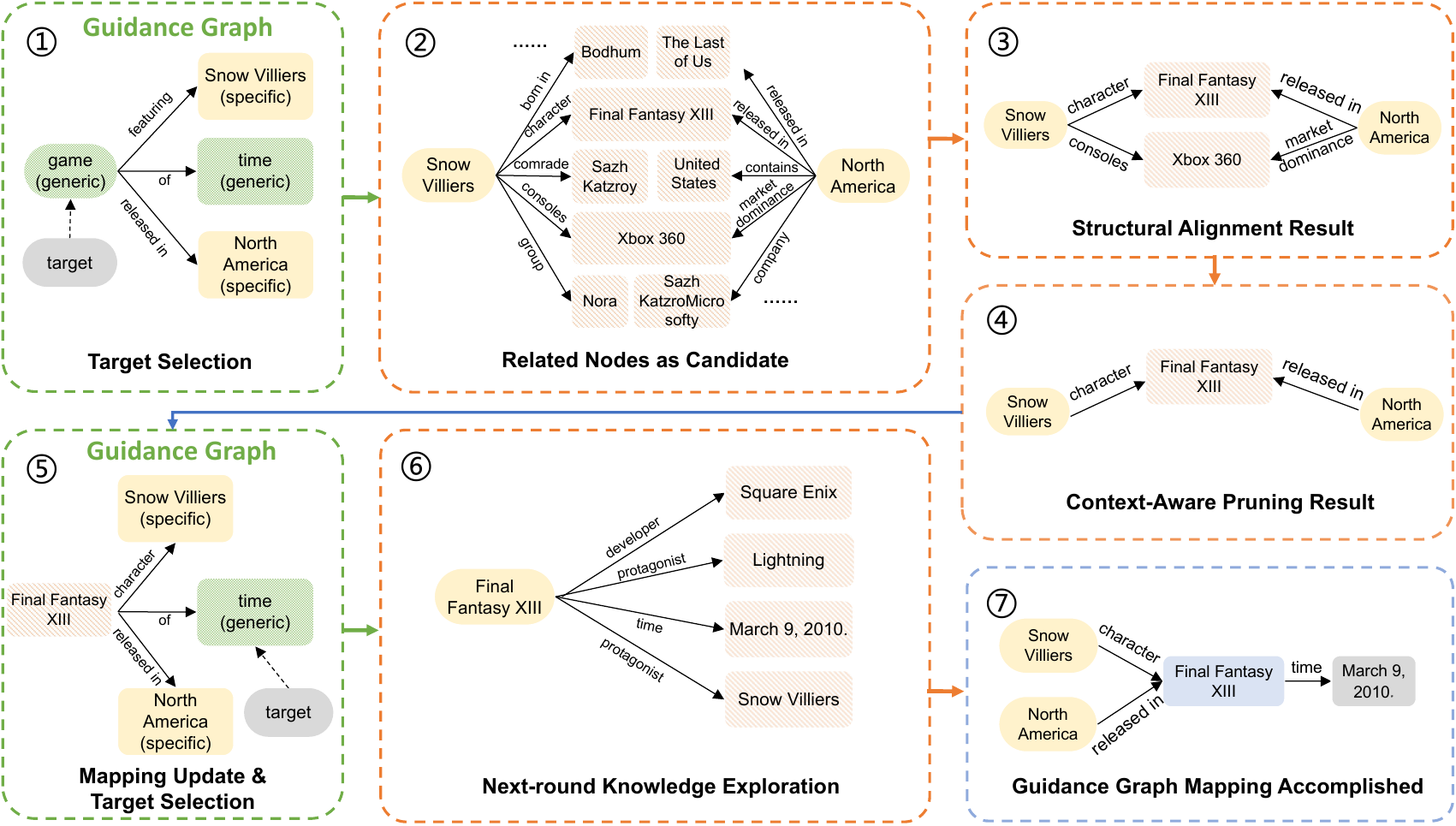}
    \caption{Iterative knowledge exploration maps Guidance Graph nodes and edges to the knowledge graph, finding structurally and semantically matching subgraphs.}
    \label{fig:ke}
\end{figure*}
\subsection{Knowledge Exploration}
\label{Knowledge Exploration}
To avoid confusion with entities and relations in the knowledge graph, we refer to the entities and relations in the Guidance Graph as clue entities and clue relations. We progressively maps clue entities and clue relations to the knowledge graph through iterative knowledge exploration, ultimately identifying knowledge subgraphs that exhibit both structural and semantic alignment with the Guidance Graph. The prompts used are shown in Appendix.

\subsubsection{Starting Points}
Knowledge exploration begins from starting points and gradually expands outward. Question-guided approaches tend to treat these starting points as prior knowledge, while clue-guided methods attempt to use all fine-grained clues as starting points. The former lacks practicality, whereas the latter may mistakenly treat conceptual entities (e.g., countries) as starting points, directly leading to exploration failures. 

In our approach, we select specific keywords from the Guidance Graph as starting points. Knowledge graphs typically contain entities with the same names as these specific keywords, which can be retrieved using basic database operations. Once any starting point is retrieved, knowledge exploration can proceed without requiring the identification of all specific keywords. In the experimental section, we investigate the impact of starting point retrieval success on our method.
The mapping between clue entities in the Guidance Graph and their corresponding entities in the knowledge graph is maintained in $clue\_mapping=\{clue: entities\}$, which includes the retrieved starting points as part of this mapping.

\subsubsection{Target Selection for Exploration Rounds}
To ensure that the resulting knowledge subgraph aligns structurally and semantically with the Guidance Graph, we select clue entities as targets $next\_e$ from the Guidance Graph and determine their contexts $(next\_e, next\_r, current\_e)$ or $(current\_e, next\_r, next\_e)$ before each iteration of exploration. The selection criteria are as follows:

\begin{itemize}
    \item Either $(next\_e, next\_r, current\_e)$ or $(current\_e, next\_r, next\_e)$ exists in $\text{Guidance Graph}$, where $current\_e  \in \bigl\{ c \mid (c: e) \in clue\_mapping \bigr\}$
    
    \item $\text{clue\_e} \notin \bigl\{ c \mid (c, e) \in clue\_mapping \bigr\}$ 
\end{itemize}
Multiple valid target knowledge candidates may exist, and their mapping order is permutation-invariant – random selection suffices.

\subsubsection{Structural Alignment}
To ensure structural consistency between the retrieved knowledge subgraph and the Guidance Graph, we perform structural alignment on the current candidate subgraph during each knowledge exploration iteration. The candidate subgraph consists of two components: the mapped subgraph and the candidate entity set. The mapped subgraph comprises all mapped entities from $clue\_mapping=\{clue: entities\}$, while the candidate entity set $E_{cand}$ contains all entities associated with $current\_e$'s mapped entity in the knowledge graph, which can be obtained through simple database operations. 

In cases where only $current\_e$ has an edge to the target entity $next\_e$ in the Guidance Graph, the candidate subgraph naturally aligns with the Guidance Graph's structure. For illustrative purposes, we assume the Guidance Graph contains an additional clue entity $related\_e$ that also connects to $next\_e$ alongside $current\_e$.

We retrieve two sets of mapped entities from $clue\_mapping=\{clue: entities\}$: the entity set $E_{curr}$ corresponding to $current\_e$ and the entity set $E_{rel}$ corresponding to $related\_e$.

For each entity in the candidate set $E_{cand}$, we consider it structurally valid if it satisfies both connectivity conditions in the knowledge graph: (1) having an edge with at least one entity from set $E_{curr}$ and (2) having an edge with at least one entity from set $E_{rel}$. This dual-connectivity requirement ensures that selected candidates simultaneously maintain structural relationships with both clue entities as specified in the Guidance Graph. The structurally valid candidate entity set and the already mapped subgraph form a structurally valid candidate subgraph.

Upon completing this round of knowledge exploration and obtaining new mappings \{next\_e: new\_entities\}, a holistic structural alignment is performed. This alignment propagates outward from the newly discovered mappings to encompass all existing mappings.
For any entity in entity set $E_{curr}$ or $E_{rel}$ that shares no edges with entities in new\_entities, that entity is removed from the mapping set. The same connectivity check is then applied to other clue entities connected to $current\_e$ and $related\_e$, and all subsequently completed mappings.
This process ensures mutual connectivity among all mappings, which maintains the overall connected structure of the Guidance Graph.

\begin{table*}[htbp]
    \centering

\begin{tabular}{l|cc|cc}
        \hline
       \multicolumn{1}{c|}{\multirow{2}{*}{\textbf{Method}}}& \multicolumn{2}{c}{\textbf{WebQSP}} & \multicolumn{2}{c}{\textbf{CWQ}}    \\
        \cline{2-5}
        & \textbf{partial match}    & \textbf{complete match}     & \textbf{partial match}      & \textbf{complete match}   \\
        \hline
        \multicolumn{5}{c}{\textit{without external knowledge}} \\
        \hline
IO prompt \cite{brown2020language} w/DeepSeek-V3 & 63.3  &  35.3  &  44.8 &  44.8    \\
COT \cite{wei2022chain} w/DeepSeek-V3   & 70.5    &  41.4   &   46.7 &    46.7   \\ 
        \hline
        \multicolumn{5}{c}{\textit{with external knowledge}} \\
        \hline

ToG \cite{sun2024think} w/Llama3-8B-Instruct  &  55.6   &   32.3   &  -    &  -    \\
PoG \cite{chen2024plan} w/Llama3-8B-Instruct  &   63.4 &  34.4  &    -    &  -  \\
FiSKE \cite{TAO2025114011} w/Llama3-8B-Instruct &    70.8 &  40.4    &   -     &  -     \\
\hline
StructGPT \cite{jiang2023structgpt} w/GPT-3.5     &  72.6 &  -   & 54.3  &    54.3   \\

KB-BINDER \cite{li2023few} w/GPT-3.5     &  74.4  &   -  &  -     &    -   \\
ToG \cite{sun2024think} w/GPT-3.5  &  76.2   &   -   &  57.1    &  57.1  \\
        \hline
PoG \cite{chen2024plan} w/DeepSeek-V3  &   81.9 &  60.7  &    55.7    &  55.7  \\

FiSKE \cite{TAO2025114011} w/DeepSeek-V3 &   \textbf{82.5} &  61.1    &    50.2     &  50.2     \\

\hline

Ours w/Llama3-8B-Instruct &    79.3 &  54.1    &    56.7    &  56.7     \\
Ours w/DeepSeek-V3 &    81.8 &  \textbf{64.5}    &    \textbf{71.8}    &  \textbf{71.8}    \\

        \hline
\end{tabular}
\caption{Results for WebQSP and CWQ. In CWQ, each question has only one answer, so partial matching is equivalent to complete matching.}
\label{table:freebase}
\end{table*}

\subsubsection{Context-Aware Pruning}

The structurally aligned candidate entity set may still contain irrelevant entities. To address this, we design a context-aware pruning method to refine the candidates, ensuring the selected entities maintain consistent contextual semantics with the Guidance Graph.

The candidate entity set could be extremely large, making it computationally expensive to individually evaluate the context of each candidate entity. Since the number of relations in a knowledge graph is typically much smaller than the number of entities, we instead operate on a candidate relation set $R$. This set consists of all relations between the candidate entity set and entity set $E_{curr}$.

For the context $(next\_e, next\_r, current\_e)$ or $(current\_e, next\_r, next\_e)$ of the target determined in target selection phase, we construct context phrase through sequential concatenation of their components. The LLM is then instructed to select the most appropriate relation $r \in R$ that can substitute $next\_r$ while preserving the original semantic meaning of the target phrase. This substitution approach ensures optimal semantic matching with $next\_r$ while maintaining contextual entity semantics.

Upon obtaining the mapped relation $r$, we define $E_{\text{mapping}}$ as the set of entities in the candidate entity set that are  connected to $r$ and update $clue\_mapping$ with the new mapping $next\_e: E_{\text{mapping}}$. We then perform the aforementioned holistic structure alignment to remove disconnected entities from the $clue\_mapping$. 

Our pruning method offers two key advantages over existing relation pruning approaches: target-awareness and context-awareness, making it theoretically more precise. The proposed structure alignment method not only ensures structural consistency between the explored subgraph and the Guidance Graph, but also completely supersedes the role of conventional entity pruning methods, as irrelevant entities are automatically eliminated during the holistic structure alignment process.

\begin{table*}[htbp]

    \centering

    \begin{tabular}{l|cc|cc}
        \hline
         \multicolumn{1}{c|}{\multirow{2}{*}{\textbf{Method}}} & \multicolumn{2}{c}{\textbf{1-Hop Query}}& \multicolumn{2}{c}{\textbf{multi-Hop Query}} \\
        \cline{2-5}
         &  \textbf{partial match} & \textbf{complete match} & \textbf{partial match} & \textbf{complete match} \\
            \hline
        IO prompt \cite{brown2020language} w/DeepSeek-V3      & 35.8 & 22.1 & 12.4 & 6.6 \\
        CoT \cite{wei2022chain} w/DeepSeek-V3      & 37.2 & 23.3 & 12.8 & 7.1\\
        \hline
        PoG \cite{sun2024think} w/DeepSeek-V3 &67.7 & 62.6& 23.9& 13.0\\
        FiSKE \cite{TAO2025114011} w/DeepSeek-V3  & 71.1 & 62.8 &28.3 & 12.9 \\
        Ours w/DeepSeek-V3   & \textbf{89.9} & \textbf{75.8} &\textbf{81.8} & \textbf{56.8}\\
        \hline
        
    \end{tabular}
    \caption{Results for the agricultural graph.}
    
    \label{table:agr}
\end{table*}

\begin{table*}[htbp]
\centering

\begin{tabular}{lccccc}
\toprule
\textbf{Dataset} & \textbf{Method} & \textbf{LLM Call} & \multicolumn{1}{l}{\textbf{Input Token}} & \multicolumn{1}{l}{\textbf{Output Token}} & \multicolumn{1}{l}{\textbf{Total Token}}  \\
\midrule

\multirow{3}{*}{WebQSP} & PoG & 11.3 & 6590.2 & 427.0 & 7017.2  \\
 & FiSKE & 7.9 & 3079.0 & 1379.7 & 4458.7  \\
 & \textbf{Ours} & 8.6& 3264.5 & 584.0 & 3848.5  \\

\midrule

\multirow{3}{*}{CWQ} & PoG & 23.4 & 15483.4 & 694.8 & 16178.2 \\
 & FiSKE & 9.4 & 3578.8 & 1828.7 & 5407.5  \\
 & \textbf{Ours} & 10.2  & 4052.0 & 708.6 & 4760.6  \\
 
\midrule

\multirow{3}{*}{agr-one-hop} & PoG & 5.6 & 2233.9 & 217.4 & 2451.4 \\
& FiSKE & 5.9  & - & - & - \\
 & \textbf{Ours} & 4.6  & 1000.0 & 294.4 & 1294.4  \\

\midrule

\multirow{3}{*}{agr-multi-hop} & PoG & 6.9 & 3243.6 & 379.6 & 3623.3 \\
& FiSKE & 13.1  & - & - & - \\
 & \textbf{Ours} & 4.1  & 975.5 & 380.7 & 1356.2  \\

\bottomrule
\end{tabular}
\caption{Efficiency Comparison Across Datasets. The agr denotes the agricultural dataset. }
\label{tab:llm_performance}
\end{table*}

\subsection{Dynamic Branch Selection under Knowledge Incompleteness}

Knowledge graphs typically exhibit incompleteness, where not all specific-keyword-derived clue entities can be mapped to corresponding nodes, or their associated relations may remain unmappable. When direct-constraint pruning fails despite identifying $\text{related\_e}$ due to missing relations, we generate semantic phrases from $(\text{current\_e}, \text{next\_r}, \text{next\_e})$ and $(\text{related\_e}, \text{related\_r}, \text{next\_e})$. The LLM selects the phrase that better matches the query context. Selection of $\text{related\_e}$ triggers new exploration starting from it, while the starting point of non-selected branches are permanently pruned from subsequent exploration paths.

\begin{table*}[htb]
\centering
\begin{tabular}{lccccccc}
\toprule
\textbf{Dataset} & \textbf{Method} & \textbf{LLM Call} & \textbf{Input Token} & \textbf{Output Token} & \textbf{Total Token} & \textbf{Partial match} & \textbf{Complete match} \\ 
\midrule
CWQ & \textbf{Ours} & 10.2 & 4052.0 & 708.6 & 4760.6 & 0.7179 & 0.7179 \\
Filtered-CWQ & \textbf{Ours} & 12.0 & 4875.4 & 769.9 & 5645.3 & 0.7460 & 0.7460 \\
\bottomrule
\end{tabular}
\caption{Experimental results regarding starting points. In Filtered-CWQ, questions without identifiable starting points were removed. In our approach, such questions are directly answered by the LLM.}
\end{table*}

\begin{table*}[htbp]
\centering

\begin{tabular}{lccccc}
\toprule
\textbf{Variant} & \textbf{partial match}& \textbf{LLM Call}& \textbf{Input Token} & \textbf{Output Token} & \textbf{Total Token} \\
\midrule
Ours       & 71.8 &  10.2 &  4052.0  & 708.6 & 4760.6        \\
\hline
w/o Context-Aware Pruning            & 53.8 &  7.9 &  2908.5  & 591.1 & 3499.6   \\
w/o Structural Alignment   & 67.9 & 11.6  & 4761.9 & 829.5 & 5591.4     \\
w/o Dynamic Branch Selection   & 69.2 &  10.0 & 3983.8  & 701.4 & 4685.2      \\

\bottomrule
\end{tabular}
\caption{Ablation Studies on CWQ set.}
\label{tab:ablation}
\end{table*}

\section{Experiments}
\subsection{Experimental Settings}

\textbf{Datasets and Evaluation Metrics.} To evaluate the performance of our proposed paradigm, We selected one open-source knowledge graph and one self-constructed graph as external knowledge bases for experimentation. The open-source knowledge graph is Freebase \cite{bollacker2008freebase}. The graph that we constructed ourselves will be publicly available soon and is referred to in this paper as the agricultural knowledge graph.

Freebase is a large-scale, semi-structured database supported by Google, designed to collect and connect information about millions of entities and their relationships worldwide. With its exceptionally large volume of relations and entities, Freebase perfectly fits our knowledge-dense scenario. We conduct experiments on two QA sets with Freebase as external knowledge base: WebQSP \cite{yih2016value} and CWQ \cite{cwq}.  

The agricultural knowledge graph that we constructed includes over 100,000 entities and 1 million triples. Although the number of entities is relatively modest compared to general-purpose knowledge graphs like Freebase, its relational structure remains highly complex. On this basis, we built single-hop and multi-hop question-answering sets.

We evaluate multi-answer questions through two distinct metrics: (1) partial match, where success requires retrieval of at least one correct answer, and (2) complete match, which necessitates identification of all correct answers.

\textbf{Baselines.} We compared our approach with five baseline methods: standard prompting (IO prompt) \cite{brown2020language}, chain of thought prompting (CoT prompt) \cite{wei2022chain}, ToG \cite{sun2024think}, PoG \cite{chen2024plan}, StructGPT \cite{jiang2023structgpt}, KB-BINDER \cite{li2023few} and FiSKE \cite{TAO2025114011}. IO prompt and CoT prompt are two knowledge-free methods, used to measure how many questions LLMs can answer solely based on their internal knowledge. ToG, PoG, StructGPT and KB-BINDER represent previous state-of-the-art approaches in knowledge base question answering, serving as baselines to evaluate our knowledge retrieval method's effectiveness.

\textbf{Experiment Details.} To ensure the reliability and reproducibility of the experiments, we set the temperature parameter to 0 for all LLMs. For the English QA datasets (WebQSP and CWQ), we employed the original prompts provided in the baseline method's codebase. For the Chinese agricultural QA dataset, we adapted Chinese versions of the prompts by translating the English template while preserving the structural format for each baseline method.

\subsection{Performance Comparison}
The results of WebQSP and CWQ are shown in Table \ref{table:freebase}.
On the WebQSP dataset, our approach with Llama3-8B-Instruct significantly outperforms existing methods using the same model, achieving results comparable to or even surpassing those obtained by large-scale models like GPT-3.5 or DeepSeek-V3 \cite{DeepSeekAI2024DeepSeekV3TR}. This not only demonstrates the effectiveness of our design but also highlights its particular suitability for smaller LLMs, indicating substantial practical value. When evaluated on DeepSeek-V3, our method achieves partial match performance comparable to the state-of-the-art FiSKE and PoG methods while outperforming all other approaches, and exceeds all methods in complete match results. This suggests our approach tends to retrieve more comprehensive knowledge.

Compared to WebQSP, the CWQ dataset contains more multi-hop questions. On CWQ, FiSKE underperforms PoG, validating our claim that relying solely on fine-grained information struggles with complex problems. Our method improves upon FiSKE over 20 percentage points, demonstrating its strong capability in knowledge-intensive scenarios for solving complex problems.

The experimental results on the agricultural dataset are presented in the Table \ref{table:agr}. For fair comparison, we equipped PoG with the same fine-grained information extraction method as ours, since its original implementation relies on dataset-provided topic entities which are unavailable in agricultural QA sets.
Our method demonstrates significant improvements over existing approaches, particularly for multi-hop question answering, where the advantage reaches remarkable levels. A notable observation reveals that while existing methods show substantial performance gaps between single-hop and multi-hop QA tasks, our approach consistently maintains high performance across both scenarios. This demonstrates our method's superior capability in handling complex problems within knowledge-intensive graphs, as also evidenced in Table \ref{table:freebase}.

Overall, our method demonstrates superior performance compared to existing approaches, particularly excelling in knowledge-intensive scenarios for solving complex problems while maintaining strong compatibility with smaller-scale LLMs.

\subsection{Computational Cost}

We conducted an efficiency study of our method, with results shown in the Table \ref{tab:llm_performance}. Overall, our analysis reveals that our approach demonstrates comprehensive advantages over PoG in both the number of LLM calls and total token consumption. On WebQSP (dominated by single-hop questions), our method shows modest improvements, while on CWQ (containing primarily multi-hop questions), it achieves significant gains, highlighting its superior capability in handling complex problems. 

On the agricultural dataset, our method requires slightly fewer LLM calls than PoG, while the token consumption by the LLM is significantly reduced.
We attribute this to PoG's reliance on heavily constrained prompts with strict rules to govern LLM outputs, whereas our approach embeds constraints primarily within the designed strategies rather than the prompts. This distinction may explain why our method is more friendly to smaller LLMs.
FiSKE demonstrates high efficiency on WebQSP and CWQ, but incurs significant computational overhead when handling the structurally complex agricultural dataset. In contrast, our approach consistently achieves stable and superior performance across all datasets.

\subsection{Studies on starting points}
We conducted experiments to evaluate the impact of starting point retrieval success on our proposed method with results presented in Table 4. When the starting point retrieval fails, our method defaults to having the LLM answer the question directly. It should be noted that question-guided methods like ToG and PoG directly retrieve topic entities as starting points from the dataset.

The results demonstrate that bypassing questions without identifiable starting points leads to improved answer accuracy at a marginal cost to efficiency. This observation is well-justified: while our method achieves significantly higher accuracy than direct LLM responses, it naturally incurs additional computational overhead.

Overall, while the inability to identify starting points affects our method's performance, this limitation does not outweigh its substantial advantages.

\subsection{Ablation Studies}

We conducted ablation studies and the results are shown in Table \ref{tab:ablation}.
The impact of starting points was discussed in the previous subsection. We therefore conduct ablation studies on Context-Aware Pruning, Structural Alignment, and Dynamic Branch Selection. For Context-Aware Pruning, we replace the original semantically rich target phrases with single clues. For the latter two components, we completely remove them during ablation.

The ablation study demonstrates the necessity of each component, as removing any module leads to performance degradation. 
Ablating Context-Aware Pruning causes significant performance drops, highlighting the crucial role of contextual semantics in LLM-based pruning.
Removing Structural Alignment reduces both accuracy and efficiency, confirming its dual benefits as designed.
Disabling Dynamic Branch Selection slightly reduces computational overhead but at the cost of performance, justifying its trade-off of minimal token consumption for substantial accuracy gains.

Another key characteristic of our method is its sequential nature, specifically manifested in the target selection process. The three modules (Context-Aware Pruning, Structural Alignment, and Dynamic Branch Selection) can only function effectively within this sequential framework. The demonstrated effectiveness of these modules collectively proves the necessity of maintaining this ordered execution flow.

\section{Conclusion}
In this paper, we propose Guidance-Graph-guided Knowledge Exploration, a novel paradigm that bridges the gap between unstructured queries and structured knowledge retrieval.
By introducing the Guidance Graph, our method resolves the dilemma faced by existing approaches that either suffer from redundant exploration or struggle to distinguish complex structures. Our method extracts multi-dimensional information from questions and establishes systematic construction rules for Guidance Graph, endowing it with both rich structural and semantic information.
We propose Context-Aware Pruning and Structural Alignment to efficiently and accurately identify knowledge subgraphs that maintain both structural and contextual semantic consistency with the Guidance Graph. Extensive experiments demonstrate that our method achieves significant improvements across multiple datasets, particularly demonstrating exceptional capability in distinguishing complex structures within knowledge-intensive scenarios. The results further reveal our method's strong compatibility with smaller-scale LLMs, highlighting its practical value. Comprehensive ablation studies validate the contribution of each proposed module.

\bibliography{aaai2026}

\end{document}